\pdfoutput=1

\documentclass[11pt]{article}

\usepackage{ACL2023}

\usepackage{times}
\usepackage{latexsym}

\usepackage[T1]{fontenc}

\usepackage[utf8]{inputenc}

\usepackage{microtype}

\usepackage{inconsolata}
\usepackage{graphicx}
\usepackage{booktabs}
\usepackage{hyperref}

\title{LASER: LLM Agent with State-Space Exploration for Web Navigation}

\author{Kaixin Ma \quad Hongming Zhang \quad Hongwei Wang \quad Xiaoman Pan \quad Wenhao Yu \quad Dong Yu \\
    Tencent AI Lab, Bellevue, WA \\ 
    \tt \{kaixinma,hongmingzhang,hongweiw,xiaomanpan,wenhaowyu,dyu@global.tencent.com\}}

\begin{document}
\maketitle
\begin{abstract}
Large language models (LLMs) have been successfully adapted for interactive decision-making tasks like web navigation. While achieving decent performance, previous methods implicitly assume a forward-only execution mode for the model, where they only provide oracle trajectories as in-context examples to guide the model on how to reason in the environment.
Consequently, the model could not handle more challenging scenarios not covered in the in-context examples, e.g., mistakes, leading to sub-optimal performance. To address this issue, we propose to model the interactive task as state space exploration, where the LLM agent transitions among a pre-defined set of states by performing actions to complete the task. This formulation enables flexible backtracking, allowing the model to recover from errors easily. We evaluate our proposed \textbf{L}LM \textbf{A}gent with \textbf{S}tate-Space \textbf{E}xplo\textbf{R}ation (LASER) on both the WebShop task and \url{amazon.com}. 
Experimental results show that LASER significantly outperforms previous methods and closes the gap with human performance on the web navigation task. 
\end{abstract}

\section{Introduction}
Large language models (LLMs) such as GPT-4 \cite{openai2023gpt4} have achieved remarkable performance on a wide range of natural language understanding (NLU) tasks \cite{brown2020language,ouyang2022training,wei2023chainofthought}. Recently, they have been adapted to interactive decision-making tasks such as virtual home navigation \cite{yang2023autogpt}, text-based games \cite{lin2023swiftsage} or web-navigation \cite{yao2023react,zhou2023webarena}.  
Previous methods that utilize LLMs to solve interactive tasks often implicitly assume a forward-only execution mode for the model, where they only provide a few oracle trajectories as in-context examples to teach the model how to reason step-by-step \cite{yao2023react,sridhar2023hierarchical}. In other words, the correct action is selected at every step in those oracle trajectories. This might lead to sub-optimal performance because when the model makes an unexpected mistake at test time, it would not know how to recover from it. At the same time, including many in-context examples to cover all possible scenarios is costly or unrealistic. Moreover, previous methods assume a global action space where the model is free to take any action at any step because they either define the possible actions at the beginning of the prompt or expect the LLM to figure out the possible action from in-context examples automatically. This might further increase the task's difficulty, and the LLM may perform invalid actions in certain cases. 

\begin{figure}
    \centering
    \includegraphics[scale=0.28]{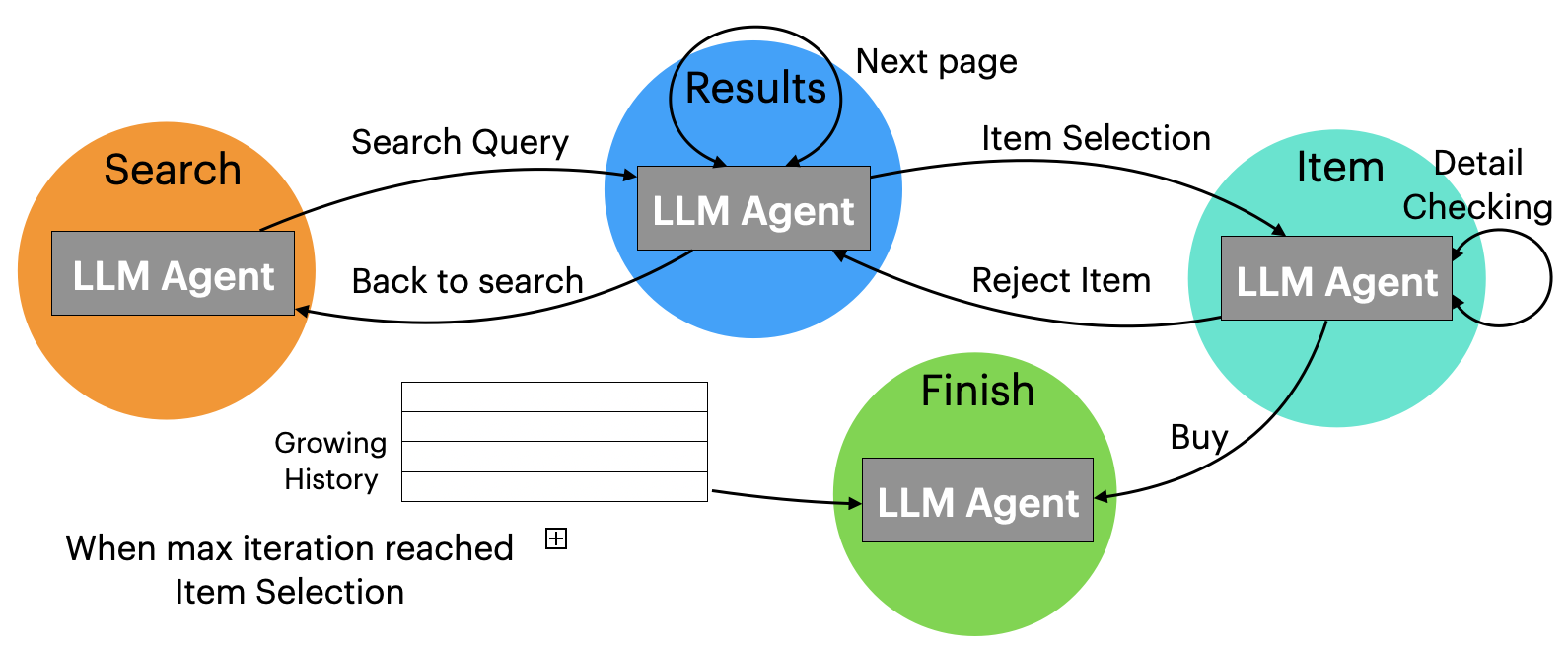}
    \caption{LASER's state transition diagram on the Webshop Task. Solid circle represent states, and the arrows represent possible state transitions. This formulation enables flexible backtracking and relieves the limitation of forward-only examples, allowing the model to better handle unfamiliar scenarios and recover from errors.}
    \label{fig:agent}
\end{figure}

To address the aforementioned issues, we propose to model the interactive tasks as state-space exploration. We first define a set of high-level possible states the LLM agent might encounter during the task execution. Then, we identify the possible action space in each state and the resulting states after performing each action. This formulation effectively converts the LLM agent's exploration in the interactive task as state transitions, where each action takes the agent from one state to another. Naturally, this allows the agent to easily recover from a wrong action: taking another action that would send it back to the previous state. Moreover, our proposed formulation associates the action space with each individual state, which reduces the task's difficulty and allows the agent to always select the valid action at any step. We evaluated our proposed LASER on the Webshop \cite{yao2023webshop} task and conducted \textit{sim-to-real} transfer experiments where we directly applied LASER to \url{amazon.com}. We show that our proposed setup enables the agent to 
complete complex user instructions without using in-context examples, and LASER significantly outperforms all previous baselines and closes the gap with human performance.  


\section{Methods}
\subsection{Problem Formulation}
Given a web environment $\mathbf{E}$ and a user instruction $\mathbf{I}$, the agent is instantiated in the environment and provided with an initial observation $\mathbf{O_0}$. The agent is expected to perform a series of actions \{$a_0$, $a_1$, ...$a_n$\} to complete the user instruction, where each $a_i$ produces a new observation $\mathbf{O_i}$ when executed in the environment. $S$ denotes the stopping state where the agent produces an output and stops exploration after reaching it. Finally, the agent's output is compared with the target to compute the metrics. 

\subsection{LLM Agent}
As previously discussed, we would like the agent to be able to handle any novel situations or mistakes that might occur during execution without exhaustively describing them via a large number of in-context examples. Thus, we propose to equip LLM agents with the state-tracking capability.
 A diagram of the state transitions of our agent is shown in \autoref{fig:agent}. We start by defining a set of possible high-level states the agent might encounter in the environment (\S\ref{ssec:state}). The LLM agent takes the user input as the overall goal and is initialized in the starting state. At every step, the agent receives state-specific system instruction, current observation, a set of permissible actions in the current states, and the history of past thoughts and actions as inputs. Then, it selects one of the actions as output, which either transitions the agent to a different state or remains in the same state (\S\ref{ssec:action}). The agent repeats the process until the stopping state or the maximum step is reached. 

Notice that with our formulation, we can provide detailed instructions to inform the agent of the possible situations in every state and how to handle them. For example, as shown in \autoref{fig:agent}, at the results state, the current results may or may not be good enough, and we instruct the agent to either select an item, go to the next page, or go back to search depending on its judgment. Hence, these instructions can be very informative to guide the agent while being much more efficient than in-context examples. Next, we describe in detail how we design the state and action spaces. 

\subsection{State Description}
\label{ssec:state}
In our work, we use the term \textit{state} to describe the current environment the agent is in, and we consider an agent to be in two different states only if the \textit{structure} of the current environment observation is different. 
This allows us to define only a handful of states to support an agent's exploration in a complex environment fully. 

After manually categorizing all possible states in the interactive task, for each state, we write a generic instruction that describes the state in detail. Specifically, we provide a sample layout of the observation the agent would receive in that state and replace all specifications in the layout with placeholders. We also provide a high-level goal and detailed instructions to act in that state. The sample layout combined with state-specific instructions allows us to inform the agent of possible observations it might receive and how to act accordingly. Therefore we no longer need to provide in-context examples to guide the agent. For the WebShop task, we define a total of four states, and the full prompts for search, results, and item states can be found in \autoref{tab:search_prompt}, \autoref{tab:result_prompt} and \autoref{tab:item_prompt} in the appendix. 

\subsection{Action Space}
\label{ssec:action}
Previous methods often implicitly assume a global action space for the model, i.e. the model is free to take any action without further constraints. Although the LLM is able to figure out valid actions to take most of the time, it might still attempt to take invalid actions in certain cases. Thus after defining all possible states for the task, we further identify the action space for each state to rule out such possibilities. Specifically, we define a set of permissible actions that the agent can choose from for each state, which ensures that the agent always performs valid actions. The state-action mapping for our agent is shown in \autoref{tab:state_action} in the appendix. In practice, permissible actions can also be determined heuristically, e.g., identifying all clickable buttons on a webpage. 

Inspired by the ReAct method \cite{yao2023react}, we also ask the agent to produce a thought at every step and then select an action based on its thought. The agent keeps repeating the thought-and-action process until it reaches the stopping state or the maximum step is reached. We also define a memory buffer to store the intermediate results (the items examined but considered non-matching) during the exploration. This is similar to human behavior in that people typically find a few backup options before finding the desired item. When the agent is forced to stop after the maximum number of steps, it selects one of the intermediate results as the final output, and we call this the backup strategy. 

\begin{table}[t!]
\centering
\resizebox{\linewidth}{!}{
\begin{tabular}{lcc}
\toprule
 & Success Rate & Reward\\
\midrule
ASH \cite{sridhar2023hierarchical} & 30.2 & 56.7 \\
ReAct \cite{yao2023react}* & 40.0 & 66.6 \\
ReAct (ours rerun) & 34.0 & 59.7 \\
WebGUM \cite{furuta2023multimodal} & 45.0 & 67.5 \\
\hline
LASER - backup & 48.4 & 71.2 \\
LASER & \bf 50.0 & \bf 75.6 \\
\hline
Human Expert \cite{yao2023webshop} & 59.6 & 82.1 \\
\bottomrule
\end{tabular}
}
\vspace{-0.05in}
\caption{Results on WebShop Task. 
*simplified setting 
}
\label{tab:zero-shot}
\end{table}

\section{Experiments}
We conduct our experiments on the WebShop task \cite{yao2023webshop}. We used 500 test set instructions for evaluation and adopted reward and success rate as metrics following previous works \cite{yao2023webshop}. We used GPT-4-0613 to power LASER and its function-calling ability to implement action selection step. 
We compare against the following baselines: ReAct \cite{yao2023react} is a prompting method designed for interactive decision-making tasks. At every step, the LLM agent receives an observation and can either produce a thought or an action. The agent accumulates all of the past observations, thoughts, and actions in its prompt, using a full trajectory of exploration as an in-context example. The original ReAct uses PaLM \cite{chowdhery2022palm} as its LLM backbone. To make a fair comparison, we also rerun the ReAct method with GPT-4-0613. ASH \cite{sridhar2023hierarchical} builds on top of ReAct and adds a summarization step that condenses the agent observation and acts based on the condensed information. WebGUM \cite{furuta2023multimodal} is a supervised method that finetunes FlanT5-XL model \cite{chung2022scaling} on 1K human demonstrations provided by the WebShop task. Moreoever, we experimented with \textit{sim-to-real} transfer experiments where we directly apply LASER to \url{amazon.com} without modification. We follow the same settings as \citet{yao2023webshop} and evaluated on 100 test set instructions and then manually evaluated results. More detailed experimental setup is discussed in \autoref{appendix:exp}.

\begin{table}[t!]
\centering
\resizebox{\linewidth}{!}{
\begin{tabular}{lccccc}
\toprule
 & SR & Reward & Att. & Opt. & Type. \\
\midrule
LASER & 62.0 & 85.4 & 85.5 & 75.0 & 97.0\\
Human \cite{yao2023webshop} & 65.0 & 88.2 & 86.2 & 76.3 & 99.0 \\
\bottomrule
\end{tabular}
}
\vspace{-0.05in}
\caption{Results on Amazon.com.
}
\label{tab:amazon}
\end{table}

\begin{table}[t!]
\centering
\resizebox{\linewidth}{!}{
\begin{tabular}{lcc}
\toprule
 & Success Rate & Reward\\
\midrule
LASER & \bf 52.0 & \bf 77.6 \\
LASER + One-shot & 50.0 & 74.9 \\ 
LASER - function call & 50.0 & 76.2 \\ 
LASER (text-davinci-003) & 38.5 & 70.2 \\ 
\bottomrule
\end{tabular}
}
\vspace{-0.05in}
\caption{Ablation Results on the WebShop Task. The standard LASER is powered by GPT-4 under zero-shot. 
}
\label{tab:ablations}
\end{table}

\section{Results}
The overall results of our experiments are shown in \autoref{tab:zero-shot}. Our early experiments showed that the ReAct agent often produces invalid actions. For example, when it selects an item that doesn't match the instruction, it tries to click the next page button (which does not exist) before backing to the results page. Also, the ReAct agent often got stuck in a certain action and failed to produce output. For example, the agent keeps going to the next page until the maximum step is reached. We added detailed instructions as the system prompt to try to address the issue. Despite our best efforts, the agent still makes invalid actions in some cases and achieves worse results than the original paper. On the other hand, LASER outperforms baselines by large margins on both metrics, showing the effectiveness of our approach. We further removed the backup strategy of LASER (the agent would receive a 0 score when the maximum budget runs out) to make a more fair comparison with ReAct. We see that our method still outperforms baselines by very large margins. The results from the transfer experiments are shown in \autoref{tab:amazon}. Again, LASER achieves very close results compared to human performance. It's also encouraging to see that LASER even achieved better performance on this realistic environment than the WebShop, which is likely due to the stronger search engine on \url{amazon.com}.

\subsection{Analysis}
\label{ssec:analysis}
We first conduct ablation studies to understand the important design decisions of our agent. 

\noindent \textbf{Zero-shot vs Few-shot} We used state-specific instructions only to guide our agent's exploration in the environment, whereas previous works often adopt in-context examples. To investigate if the agent can further benefit from in-context examples, we experimented with a one-shot setting: for every prompt in LASER, we added one example input-output pair between our system instructions and current inputs, and the rest of the agent remains the same. Due to the limited computing budget, we only ran our ablation studies on 200 instructions. The results are shown in \autoref{tab:ablations}. We see that adding an in-context example actually leads to worse performance. Since LASER already performs valid actions 100\% time, we hypothesize that the agent understands the task well without in-context examples and the added example is actually distracting the agent in some cases. 

\begin{figure}
    \centering
    \includegraphics[scale=0.3]{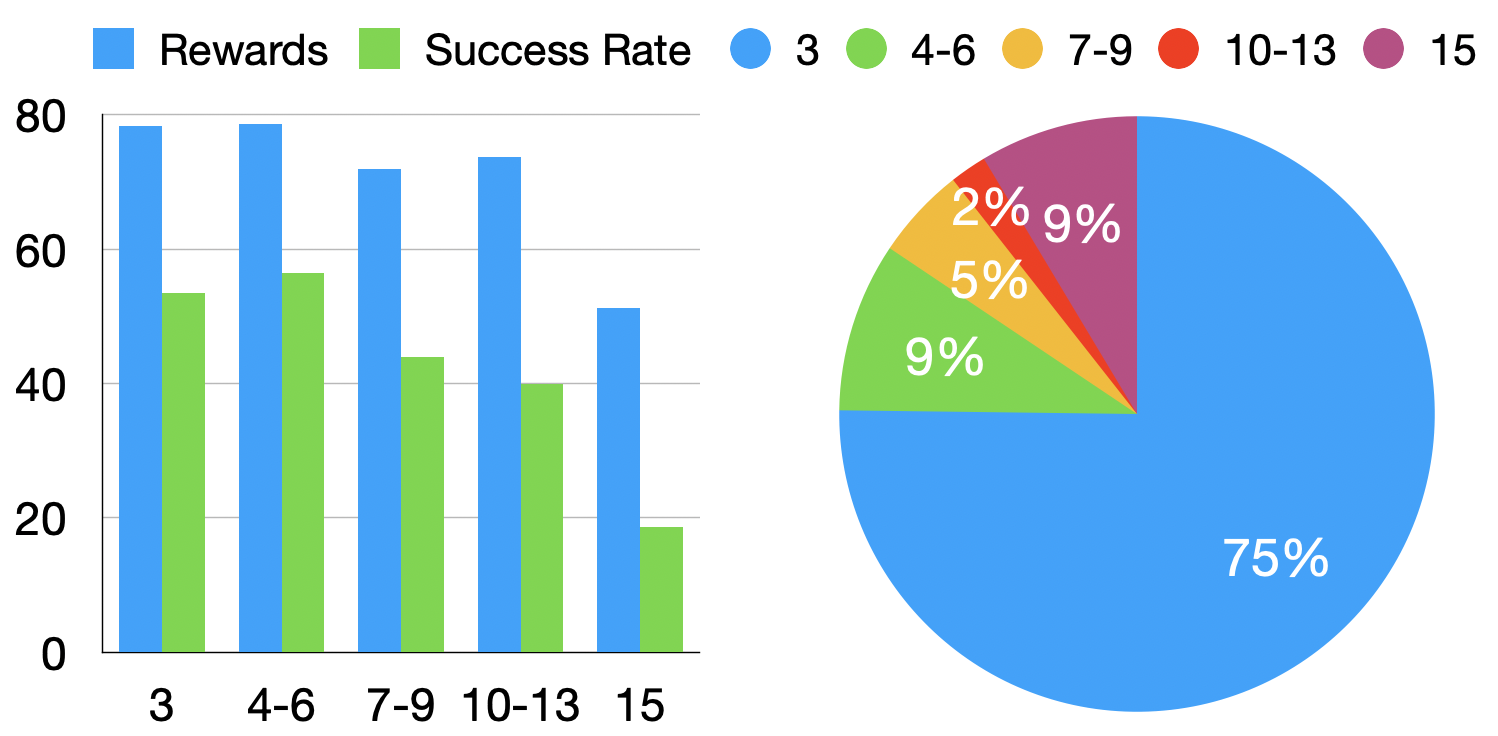}
    \caption{Left: LASER's performance for test set episodes of different lengths. Right: The distribution of the number of steps LASER takes to complete tasks} 
    \label{fig:step}
\end{figure}

\noindent \textbf{Effect of function-calling} LASER takes advantage of the function-calling functionality that is enabled only for GPT models after 06/13/23. Thus, we are interested to see the effect of replacing this design with regular text generation. To do so, instead of passing the permissible actions as a list of functions, we convert each action as a Python dictionary describing its purpose and arguments and then append them to the prompt. We then ask the LLM to generate output in JSON format to represent the action it selects with appropriate arguments. The results are shown in \autoref{tab:ablations}. Again, the agent without function calling performs slightly worse on these 200 episodes. It shows that the function calling functionality can be leveraged to boost performance when building interactive agents, suggesting a direction for building future LLMs.

\noindent \textbf{Performance vs trajectory length} Here, we are interested in seeing the length of LASER's trajectories and their effect on the overall performance. We plot the distribution of trajectory length in \autoref{fig:step} and the agent's performance for each length group. We notice that most of the time, the agent only took three state transitions to reach the finish state, which is search-select-buy. From the left figure, the agent's performance generally decreases as the trajectory gets longer. However, the drop is less significant compared to the observation made for ReAct and ASH agent \cite{sridhar2023hierarchical}, which further shows the effectiveness of our agent. Finally, for the length 15 group, 
for which the agent is forced to stop and select from the browsing history, the performance is much lower than other groups. While not surprising, it has a non-zero success rate, showing that there are cases where the agent found a matching item but failed to recognize it as the target in the first pass.

\noindent \textbf{Generalization to different LLMs} 
We adopted the text-davinci-003 model to see if LASER can transfer well to a less powerful non-chat model. Since this model does not support function-calling, we adopted the approach described earlier to prompt the model to generate JSON output to represent actions. The results are shown in \autoref{tab:ablations}. Although switching to text-davinci-003 leads to a large drop in performance, our model still achieves better results than the baselines. It shows that our proposed agent can be easily adapted to other LLMs with different capabilities. With more powerful models in the future, our agent could potentially surpass human performance on this task. We also conducted case studies to inspect the failure modes of LASER and additional results are in \autoref{sec:case}. We discuss related works in \autoref{sec:related}.

\section{Conclusions}
We proposed an LLM agent, LASER, that models interactive web navigation tasks as state-space exploration. Our formulation allows the agent to handle novel situations, easily backtrack from mistakes, and always perform valid actions. Guided solely by the state-specific instructions without any in-context examples, LASER outperforms all baselines on the WebShop task by large margins and closes the gap with human performance on the real-world shopping website. Our analysis shows that LASER is also more robust to longer trajectories and generalizes well to other LLMs.

\section*{Limitations}
In this work, we have only experimented with the task of finding the target item for the shopping domain. Despite its challenging nature, it does not cover all tasks user typiclaly do on an e-commerce website, e.g., tracking orders or checking order history. For future work, it would be interesting to enhance LASER's ability so that it can handle such popular tasks in the shopping domain. Also, it would be interesting to equip LASER with more tools such as a knowledge retriever \cite{ma-etal-2023-chain} or a calculator \cite{gao2022pal}, so that it can handle more complex instructions. 

Our LASER requires manual annotation of possible states in the environment and their corresponding descriptions. Because of this, our method might only be suitable for building agents for specific domains (rather than open-world web agents), where only a handful of states are required, e.g. e-commerce or travel booking. For future directions, we envision a hierarchical multi-agent system, in which each specific domain is governed by an agent like LASER, and a general open-world agent just collaborates with other domain agents to complete various user instructions. 

Regarding potential risks of our work, we think extra caution and testing are required before deploying LASER to real-world scenarios. When conducting experiments on the Webshop task, we allow the agent to take any action permitted in the environment because of its simulated nature. However, certain actions may have hard-to-recover consequences in the real world. For example, clicking the buy button in a real shopping site. Therefore we forced the agent to stop when it decides to buy the item when experimenting on \url{amazon.com}. In general, as LASER's success rate is still far from being perfect, it might require additional human verification before proceeding with actions that have high-stakes. 

\bibliography{anthology,custom}
\bibliographystyle{acl_natbib}
\appendix

\section{Related Works}
\label{sec:related}
Interactive decision-making tasks such as web navigation have become popular recently \cite{liu2018reinforcement,yao2023webshop,deng2023mindweb,zhou2023webarena}, while some efforts have tried to solve these tasks by finetuning pretrained language models on a large corpus of demonstration data \cite{gur2022understanding,furuta2023multimodal}, other attempted to build agents to navigate web environments solely relying on prompting LLMs \cite{yang2023autogpt}. Among the LLM-based approaches, ReAct \cite{yao2023react} and InnerMonologue  \cite{huang2022inner} equip the LLM with a thought process before producing actions. ASH \cite{sridhar2023hierarchical} and WebAgent \cite{gur2023realworld} focus on decomposing complex decision-making steps into a set of simpler steps, e.g. first summarizing the task-relevant content and then act upon it. Most similar to our work, Synapse \cite{zheng2023synapse} also proposed to use state-conditional prompts to guide the LLM's action. However, their focus is on decomposing the few-shot examples into atomic parts whereas our agent uses state-specific instructions alone without in-context examples to complete tasks. 

Another line of work focuses on the planning stage of LLM agents. \citet{kim2023language} proposed an agent RCI that generates a plan before acting, and then refines its action when encountering errors. Adaplanner \cite{sun2023adaplanner} further enhanced the planning approach by adaptively updating the plan during the agent's execution. Reflexion \cite{shinn2023reflexion} agent refines its plan and actions by taking environmental feedback through a trial-and-error fashion. These approaches are orthogonal to our work and can be potentially combined with our agent to enhance its performance. 

More recently, various works have tried to develop multi-modal agents. Pix2Act \cite{shaw2023pixels} and AppAgent \cite{zhang2023appagent} mostly replied on the screenshots as inputs for the agents to predict UI actions, wheras SEEACT \cite{zheng2024gpt4vision}, WebVoyager \cite{he2024webvoyager} and Dual-VCR \cite{kil2024dualview} leverage both screenshots and textual elements from websites to interact with the web environment. Our idea of modeling web navigation as state transitions can potentially be incorporated in those agents to further enhance their performance. 

\section{Experimental Details}
\label{appendix:exp}
The WebShop provides a simulated environment for online shopping, containing 1,181,436 items collected from Amazon shopping sites. Additionally, the task provides human-annotated instructions for purchasing certain items and their corresponding target items. We followed previous works and used the 500 test set instructions to evaluate our LASER and evaluate with rewards and success rate, where the agent is considered successful if the purchased item perfectly matches the target item, otherwise, if the purchased item partially matches the target item, the agent receives a partial reward (scale between 0-100). This partial reward is computed using the items' price, product category, hidden attributes and customization options.  

For our method, we used the GPT-4-0613 to power our LASER. We used the function-calling functionality to implement the action selection step. In particular, we write a short description for each action and then pass them as a list to the function-call argument of the LLM to let the model select from. We allow our agent to make 15 state transitions in maximum. In practice, if the agent has not reached the finish state after 13 state transitions, we force it to select from the history to ensure it does not exceed the budget.

\begin{figure}
    \centering
    \includegraphics[scale=0.3]{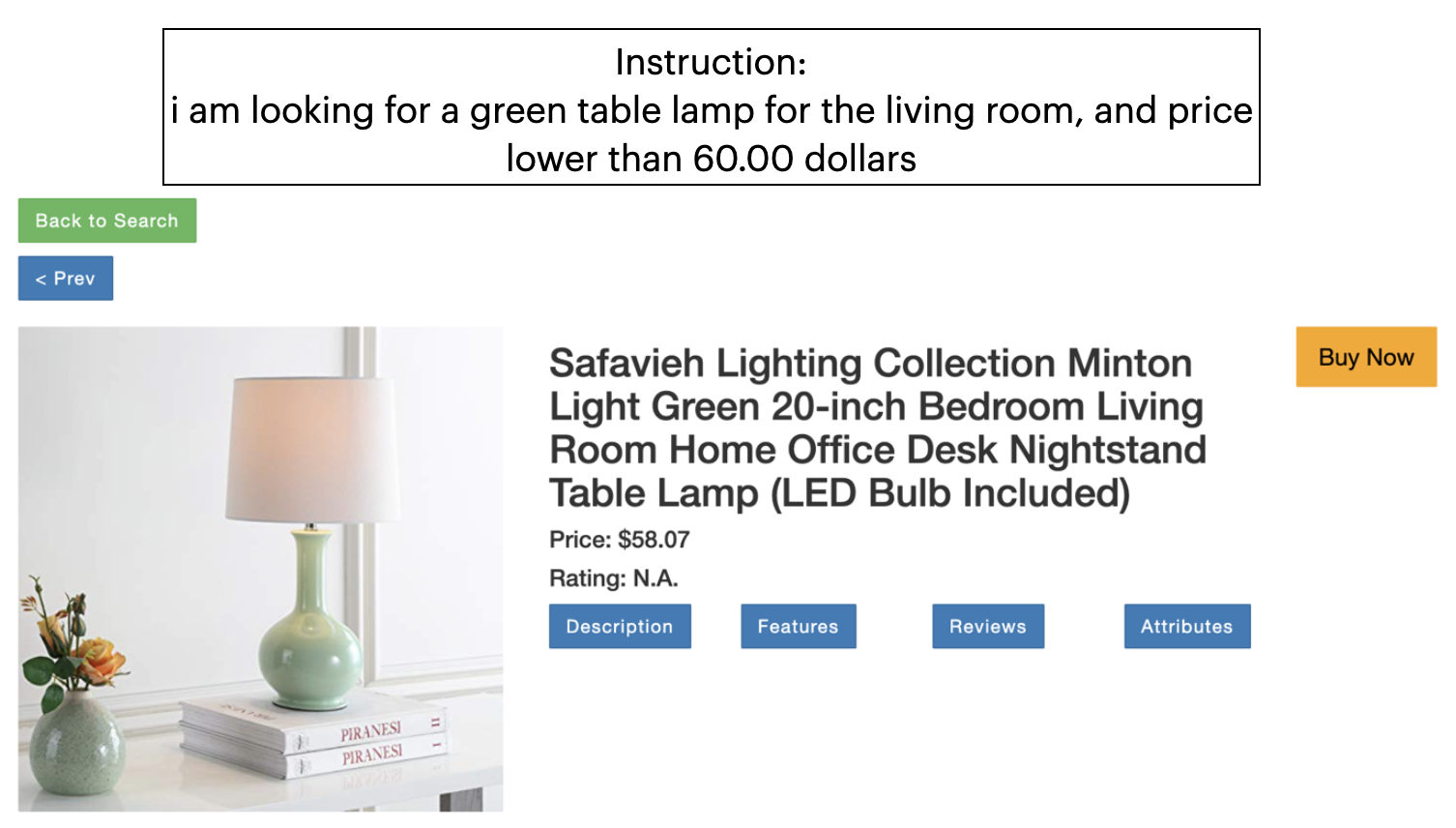}
    \caption{An example of the \textit{Item good enough} error cases, the item selected by the agent is shown and the user instruction is on the top. The reward the agent receives is 0.666.}
    \label{fig:item_good}
\end{figure}

For the \textit{sim-to-real} transfer experiments on \url{amazon.com}, we used the first 100 test set instructions from the WebShop. We following the same setting as \citet{yao2023webshop}, where we convert the webpages on \url{amazon.com} into the same format as the WebShop \footnote{\url{https://github.com/princeton-nlp/WebShop/tree/master/transfer}} then run LASER agent as is. Since we do not have the gold annotation for the items LASER selected on \url{amazon.com}, we follow \citet{yao2023webshop} and conducted human evaluation. In particular, we manually annotated item attribute matches, item option matches, item category matches and item price matches. We then computed individual reward scores as well as the overall reward score and success rate using the same functions defined for the WebShop task. Since both human and LASER achieves 100\% on item price matches, we omitted these results from \autoref{tab:amazon}.  

\begin{figure}
    \centering
    \includegraphics[scale=0.35]{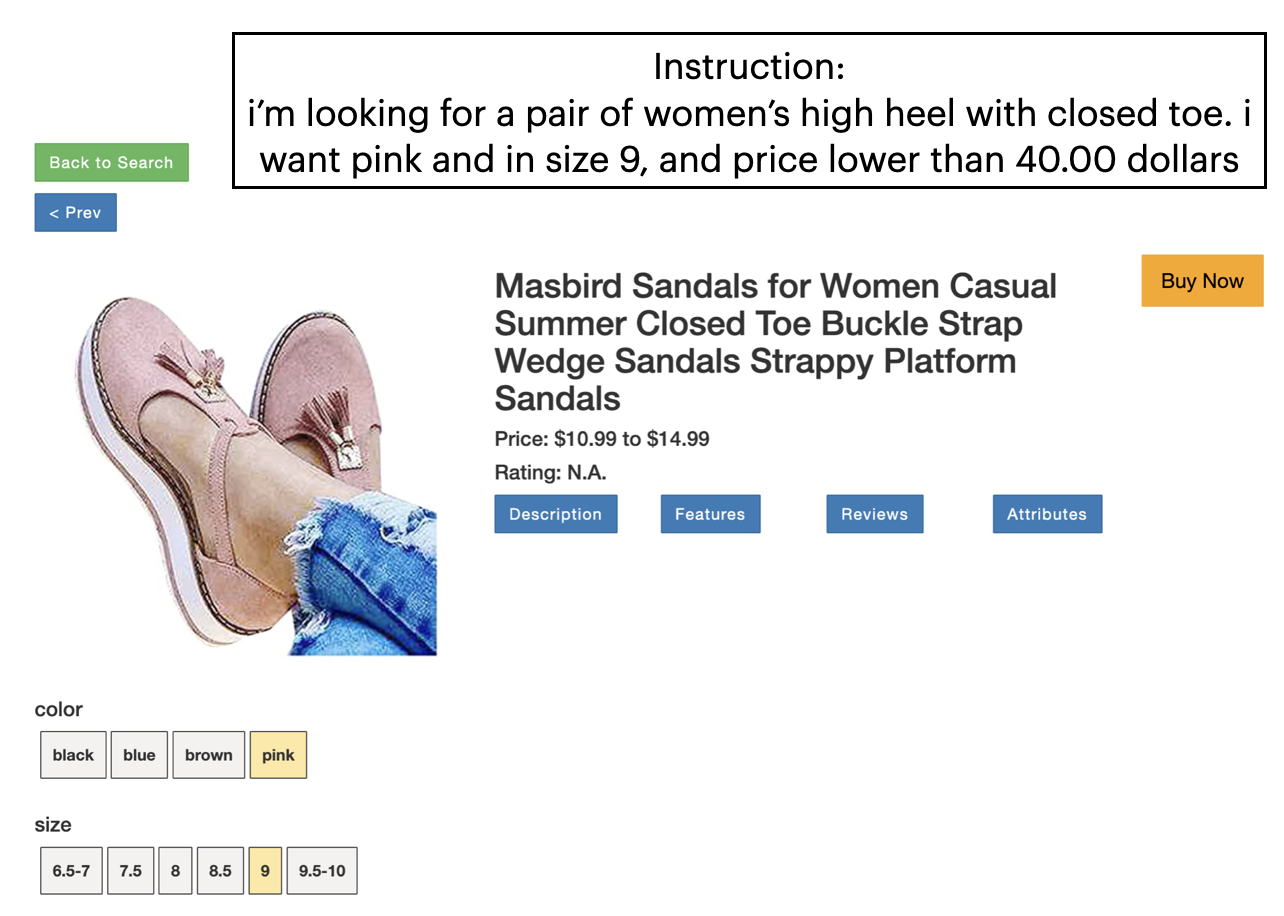}
    \caption{An example of the \textit{Missing details} error cases, the item selected by the agent is shown and the user instruction is on the top. The reward the agent receives is 0.8.}
    \label{fig:item_missing}
\end{figure}

Regarding the comparison against different baselines, we would like to note that both ReAct \cite{yao2023react} and ASH \cite{sridhar2023hierarchical} used manually written instruction and manually annotated agent trajectories as in-context demonstrations to prompt LLMs, which corresponds to our one-shot setting in \autoref{ssec:analysis}. For WebGUM \cite{furuta2023multimodal}, they used 1k human annotated gold trajectories to finetune their model. Therefore, all baselines we considered use some kind of human knowledge/prior to help the agent learn. For us, we solely relied on manual instructions to guide the LASER. 

We believe that providing high-level generalizable instructions (as done in LASER) is a more efficient ways of learning than providing low-level task-specific trajectories (e.g. WebGUM). Intuitively, the agent basically learns to abstract out some high-level insights about how to handle each scenario from the large amount of trajectories. In comparison, we can directly provide such insights to the model via a few sentences in the instruction. Taking such perspective, we can also say that the difference between our work and previous work is providing high-level generalizable human knowledge vs providing low-level case-by-case human knowledge. We believe it's desirable to provide model such high-level knowledge when it requires similar or less amount of human effort.

\section{Case Studies}
\label{sec:case}

\label{appendix:case}
We manually annotated 30 error cases from the Dev set to understand the failure cases of LASER. We broadly categorize the errors into three categories: \textit{Item good enough}: the item selected by the agent meets the user instruction from the authors' perspective but did not receive a full score. We found that 9 out of 30 cases fall into this category, and an example is shown in \autoref{fig:item_good}. The item found by the agent is indeed a green table lamp for the living room with a price within the budget, but it is considered incorrect. \textit{Retrieval failure}: none of the items returned by the search engine meets the user requirement, despite that the agent used a suitable query for retrieval. We found 12 out of 30 cases fall into this category. We hypothesize that a more effective retriever or search engine can probably address these issues. \textit{Missing details}: The item selected by the agent indeed does not match the user's instruction on certain details. We found that 9 out of 30 cases fall into this category, and an example is shown in \autoref{fig:item_missing}. In this example, although the color and size of the selected women's shoes both matched the user instructions, these are not high-heel shoes. This indicates that LASER can make mistakes when encountering items with many matching details, and it would be interesting to see if a self-feedback/verification module can address this issue \cite{madaan2023selfrefine}. 


\section{Prompts used in our experiments}
\begin{table*}[]
    \centering
    \begin{tabular}{l}
    \toprule
         You are an intelligent shopping assistant that can help users find the right item. You are given an \\
        observation of the current web navigation session, in the following format:   \\
        \\
        Current Observation: \\
        WebShop \\
        Instruction:  \\
        \{the user instruction\}   \\\relax
        [button] Search [button\_] (generate a search query based on the user instruction and select this button to \\
        find relevant items) \\
        \\
        Every button in the observation represents a possible action you can take. Based on the current \\
        observation, your task is to generate a rationale about the next action you should take. Note that if an \\
        history of  past rationales and actions is provided, you should also consider the history when generating \\
        the rationale. \\
        \bottomrule
    \end{tabular}
    \caption{The system instruction we used for the search state.}
    \label{tab:search_prompt}
\end{table*}

\begin{table*}[]
    \centering
    \begin{tabular}{l}
    \toprule
         You are an intelligent shopping assistant that can help users find the right item. You are given an \\
        observation of the current web navigation session, in the following format:   \\
        \\
        Current Observation: \\
        Instruction: \\
\{the user instruction\} \\\relax
[button] Back to Search [button\_] (select this button to go back to the search page) \\
Page {current page number} (Total results: {total number of results}) \\\relax
[button] Next > [button\_] (select this button to go to the next page of results) \\\relax
[button] \{item\_id 1\} [button\_] (select this button to view item 1's details) \\
\{name of item 1\} \\
\{price of item 1\} \\\relax
[button] \{item\_id 2\} [button\_] (select this button to view item 2's details) \\
\{name of item 2\} \\
\{price of item 2\} \\\relax
[button] \{item\_id 3\} [button\_] (select this button to view item 3's details) \\
\{name of item 3\} \\ 
\{price of item 3\} \\
\{More items...\} \\
\\
At this stage, you want to select an item that might match the user instruction. Note that even if an item\\
has non-matching details with the user instruction, it might offer different customization options to \\
allow you to match. E.g. an item may have color x in its name, but you can customize it to color y later, \\
the customization options are shown after you select the item. Thus if an item name seems relevant or \\
partially matches the instruction, you should select that item to check its details. If an item has been \\
selected before (the button has been clicked), you should not select the same item again. In other words, \\
do not select an item  with [clicked button] item\_id [clicked button\_]. Prepare your response in the \\
following format: \\
Rationale: the user wanted \{keywords of the target item\}, and we have found \{matching keywords of \\
item x\}, thus item \{item\_id x\} seems to be a match.\\
        \bottomrule
    \end{tabular}
    \caption{The system instruction we used for the result state.}
    \label{tab:result_prompt}
\end{table*}

\begin{table*}[]
    \centering
    \begin{tabular}{l}
    \toprule
         You are an intelligent shopping assistant that can help users find the right item. You are given an \\
        observation of the current web navigation session, in the following format:   \\
        \\
        Current Observation: \\
        Instruction: \\
\{the user instruction\} \\\relax
[button] Back to Search [button\_] (select this button to go back to the search page) \\\relax
[button] < Prev [button\_] (select this button to go back to the previous page of results) \\ 
\{Customization type1\}: \\\relax
  [button] {option1} [button\_]  \\\relax
  [button] {option2} [button\_] \\
\{Customization type2\}: \\\relax
  [button] {option1} [button\_]  \\\relax
  [button] {option2} [button\_] \\
\{more customization options... (if any)\} \\
\{Item name and details\} \\\relax
[button] Description [button\_] (select this button to view the full description of the item) \\\relax
[button] Features [button\_] (select this button to view the full features of the item) \\\relax
[button] Reviews [button\_] (select this button to view the full reviews of the item) \\\relax
[button] Buy Now [button\_] (select this button to buy the item) \\
\\
description: (if this is shown, the description button should not be selected again) \\
\{full description of the item (if any) or "None"\} \\
\\
features: (if this is shown, the features button should not be selected again) \\
\{full features of the item (if any) or "None"\} \\
\\
reviews: (if this is shown, the reviews button should not be selected again) \\
\{full reviews of the item (if any) or "None"\} \\
\\
Target item details (what the user is looking for): \\
keywords: \{keywords of the target item\} \\
max\_price: \{the price of the item should not exceed this\} \\
\\
At this stage, you want to verify if the item matches the user instruction. You should consider the \\
available customization options when deciding whether an item matches the user instruction. If an item \\
can be customized to match the user instruction, or if the customization options cover the user \\
specification, it is also a good match. If the item does not match the user instruction and it does not \\
provide enough customization options, you can go to previous page to view other items. You can also \\
check the item's description, features and reviews to view more details (Note that description, features \\
 and reviews could be "None", do not check  them again if they are already given). Prepare your \\
response in the following format:  \\
Rationale: the user wanted \{keywords of the target item\}, and they required the following customization \\
options: \{customization of the target item\}, the item is {keywords of the item in the current observation}, \\
and it has the following customization options: \{options available for the current item\}, which \{cover\}/ \\
\{not cover the user requirement\}, thus we should \{buy the item\}/\{check more details\}/\{go to previous \\
page to view other items\} \\
        \bottomrule
    \end{tabular}
    \caption{The system instruction we used for the item state.}
    \label{tab:item_prompt}
\end{table*}

\begin{table*}[]
    \centering
    \begin{tabular}{l}
    \toprule
         You are an intelligent shopping assistant that can help users find the right item. You are given an \\
        observation of the current environment and a rationale for the next action to be taken, \\
        in the following format:   \\
        \\
        Current Observation: \\
        \textit{The observation layout from search or result or item state, as shown from \autoref{tab:search_prompt}, \autoref{tab:result_prompt} and \autoref{tab:item_prompt}}\\
        \\
        Next action rationale: \{the rationale for the next action\} \\
\\
Your task is to perform one of the function calls based on the rationale. \\
        \bottomrule
    \end{tabular}
    \caption{The system instruction we used for generating action from thought.}
    \label{tab:action_prompt}
\end{table*}

\begin{table*}[]
    \centering
    \begin{tabular}{ll}
    \toprule
        State & Available Actions \\
        \hline
        Search & \{"name": "Search", "description": "Use this function to search for the target item in the \\
        & inventory based on keywords"\}  \\
        \hline
        Result & \{"name": "select\_item", "description": "Use this function to select one of the items from \\
        &  the search results and check its details"\}  \\
        & \{"name": "Next", "description": "Use this function to go to the next page of search results \\
        &   to view more items, if none of the items on the current page match the user instruction."\}  \\
        & \{"name": "Back\_to\_Search", "description": "Use this function to go back to the initial  \\
        &  search page. You  should use this function only if you have browsed multiple pages of   \\
        & items and checked multiple items' details in the history, and none of the items \\
        & match the user instruction."\}  \\
        \hline
        Item & \{"name": "Description", "description": "Use this function to check the description of the \\
        & item, if you are unsure if the item perfectly matches the user instruction"\}  \\
        & \{"name": "Features", "description": "Use this function to check the features of the item, \\
        &  if you are unsure if the item perfectly matches the user instruction"\}  \\
        & \{"name": "Reviews", "description": "Use this function to check the reviews of the item, \\
        & if you are unsure if the item perfectly matches the user instruction"\}  \\
        & \{"name": "Buy\_Now", "description": "Use this function to buy the current item, if the \\
        &  current item perfectly matches the user instruction."\}  \\
        & \{"name": "Prev", "description": "Use this function to go back to the results page, if the \\
        & current item does not match the user instruction "\}  \\
        \bottomrule
    \end{tabular}
    \caption{The action space of our agent in each state. Each action is implemented as a function call following the guidelines from OpenAI \footnote{\url{https://platform.openai.com/docs/guides/gpt/function-calling}}, additional parameters used in the function call are omitted here for brevity.}
    \label{tab:state_action}
\end{table*}

\section{Licenses}
The Webshop task and ReAct method are both released under MIT License. They are both released for research purposes, and our experiments are consistent with their intended usage. 

\end{document}